\documentclass[runningheads]{llncs}
\usepackage[T1]{fontenc}
\usepackage{hyperref}
\usepackage{graphicx}
\usepackage{orcidlink}
\usepackage{booktabs}
\usepackage{array}
\usepackage{subcaption}
\usepackage{multirow}
\usepackage{amsmath}
\usepackage{cleveref}
\crefname{figure}{Fig.}{Figs.}
\Crefname{figure}{Fig.}{Figs.}
\begin{document}
\title{Few-Shot Connectivity-Aware Text Line Segmentation in Historical Documents}
\titlerunning{Few-Shot Connectivity-Aware Text Line Segmentation}
%
\author{Rafael Sterzinger\orcidlink{0009-0001-0029-8463} \and
Tingyu Lin\orcidlink{0009-0008-9825-686X} \and
Robert Sablatnig\orcidlink{0000-0003-4195-1593}}
\authorrunning{R. Sterzinger et al.}
%
\institute{Computer Vision Lab, TU Wien, Vienna, AUT
\email{\{firstname.lastname\}@tuwien.ac.at}}
\maketitle              

\begin{abstract}
A foundational task for the digital analysis of documents is text line segmentation.
However, automating this process with deep learning models is challenging because it requires large, annotated datasets that are often unavailable for historical documents.
Additionally, the annotation process is a labor- and cost-intensive task that requires expert knowledge, which makes few-shot learning a promising direction for reducing data requirements.
In this work, we demonstrate that small and simple architectures, coupled with a topology-aware loss function, are more accurate and data-efficient than more complex alternatives. 
We pair a lightweight UNet++ with a connectivity-aware loss, initially developed for neuron morphology, which explicitly penalizes structural errors like line fragmentation and unintended line merges. 
To increase our limited data, we train on small patches extracted from a mere three annotated pages per manuscript.
Our methodology significantly improves upon the current state-of-the-art on the U-DIADS-TL dataset, with a 200\% increase in Recognition Accuracy and a 75\% increase in Line Intersection over Union. 
Our method also achieves an F-Measure score on par with or even exceeding that of the competition winner of the DIVA-HisDB baseline detection task, all while requiring only three annotated pages, exemplifying the efficacy of our approach. Our implementation is publicly available at: \href{https://github.com/RafaelSterzinger/acpr_few_shot_hist}{https://github.com/RafaelSterzinger/acpr\_few\_shot\_hist}.

\end{abstract}

\section{Introduction}
In historical documents, text line segmentation is a foundational task of computational document analysis.
Approaches to early layout analysis relied on heuristics, such as projection profiles and clustering of connected components~\cite{likforman1995hough,o1993document,shapiro1993handwritten}.
While effective on clean documents, these rule-based pipelines often fail when faced with the complexities inherent in historical manuscripts, such as bleed-through, elaborate ornamentation, or degraded scripts~\cite{likforman2007text}.
As a consequence of these challenges, a shift towards data-driven approaches occurred using deep neural networks such as CNNs, which re-framed the task as an end-to-end learning problem~\cite{long2015fully}.
However, the performance of deep learning models depends on the availability of large, pixel-precise annotated datasets.
In the context of historical documents, the required annotated datasets are either prohibitively expensive and time-consuming to create, or the data itself is simply unavailable.

Given this data scarcity, much research has focused on few-shot learning, where models are trained using only a handful of annotated examples.
The challenge of this low-data regime is evident in research where powerful architectures, such as DeepLabV3+\cite{chen_deeplab_2017}, achieve poor results when trained on just three pages per manuscript~\cite{zottin2025exploring}.
In response to these challenges, some have employed increasingly complex models, such as transformer-based hybrids~\cite{vadlamudi2023seamformer}.

Our approach, however, illustrates the principle of \textit{Occam's Razor}: when a simpler method outperforms more complex ones, additional complexity is not only unnecessary but can be counterproductive, as an increased number of parameters raises the risk of overfitting and reduces generalization. In this work, we make the following contributions:

\begin{itemize}
    \item We demonstrate that a simple and lightweight architecture is sufficient to significantly outperform more complex models~\cite{zottin2025exploring} in a few-shot setting.
    In addition, we incorporate an established patch-level training strategy, which efficiently handles high-resolution imagery and has proven effective in data-scarce domains~\cite{denardin2023efficient,sterzinger2024drawing} to increase data.
    
    \item 
    We propose the usage of a connectivity-aware loss function, originally proposed for neuron morphology reconstruction~\cite{grim2025efficient}, to the task of text line segmentation, to address the critical importance of structural integrity in text lines: standard pixel-level losses often fail to prevent fragmentation or unintended merging, which severely impact line-level evaluation metrics.
    
    \item We perform a thorough evaluation of our methodology on two public competition datasets, U-DIADS-TL~\cite{zottin2025exploring,icdar2025fest} and DIVA-HisDB~\cite{simistira2017compe}, under a strict few-shot constraint of only three annotated training pages.
    Our results establish a new SOTA for the former and demonstrate the approach's generalizability on the latter.
\end{itemize}

Our final model sets a new baseline on the U-DIADS-TL dataset, achieving relative gains of over 74\% in Line Intersection over Union (IoU) and nearly 200\% in Recognition Accuracy (RA) compared to the baseline reported by Zottin et al.~\cite{zottin2025exploring}.
Furthermore, on the DIVA-HisDB baseline detection task, our few-shot approach achieves performance competitive or even superior to top-ranked systems from the ICDAR 2017 competition~\cite{simistira2017compe}, despite using three instead of 20 annotated pages, i.e., 85\% less training data. 

\begin{figure}[t]
  \centering
  \begin{subfigure}{0.3\linewidth}
    \includegraphics[width=\linewidth]{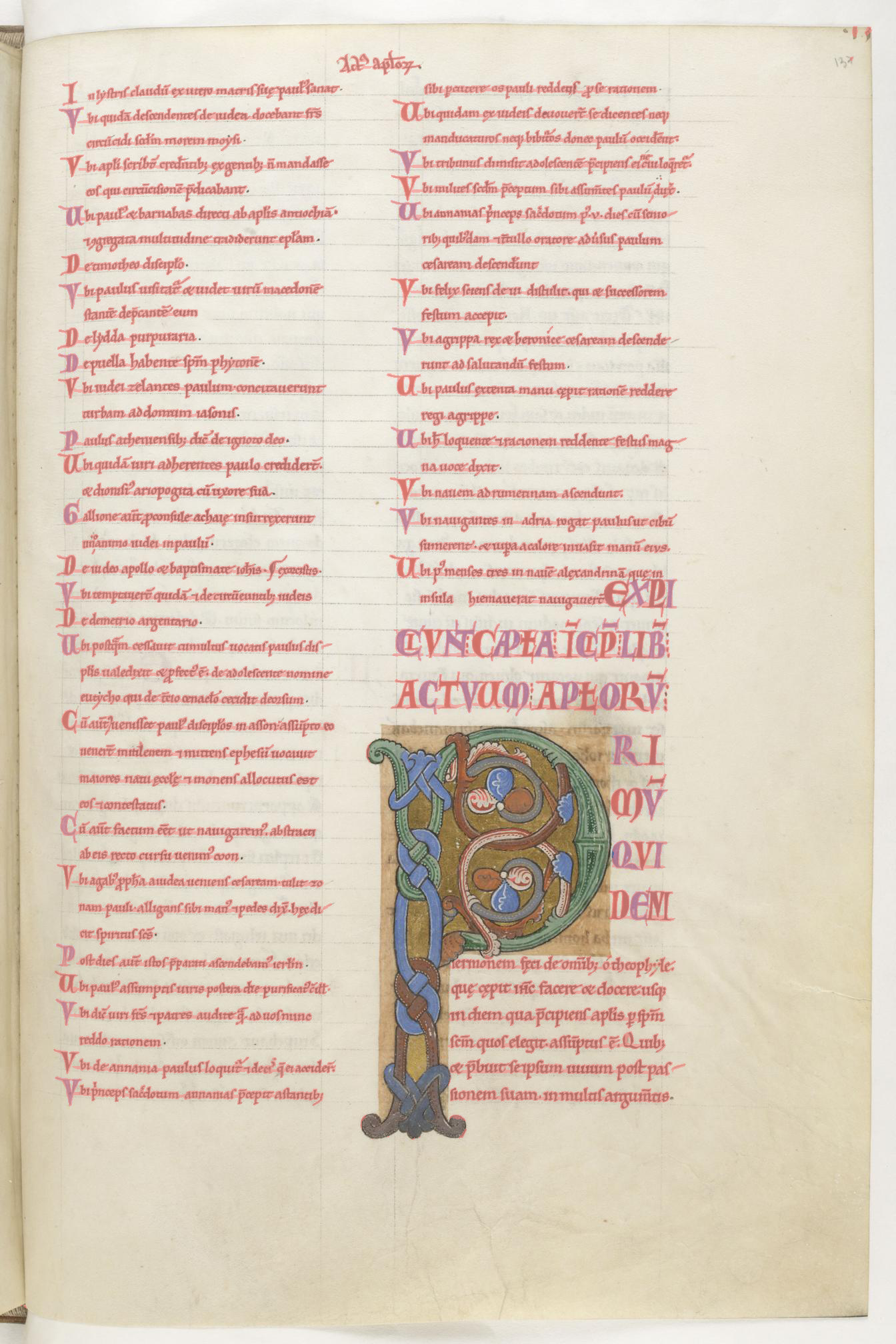}
    \caption{Latin 14396}
  \end{subfigure}
    \begin{subfigure}{0.3\linewidth}
    \includegraphics[width=\linewidth]{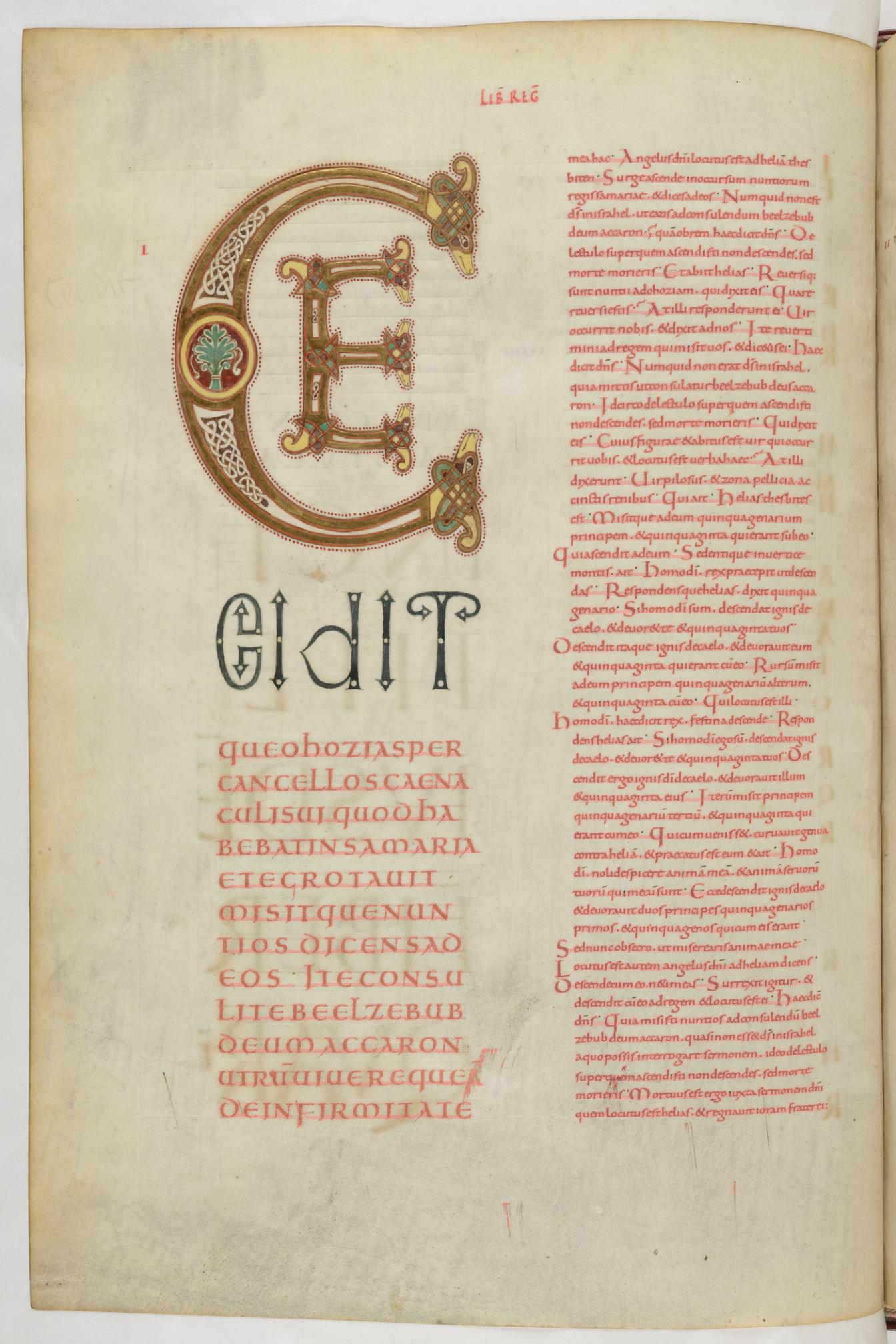}
    \caption{Latin 2}
  \end{subfigure}
    \begin{subfigure}{0.3\linewidth}
    \includegraphics[width=\linewidth]{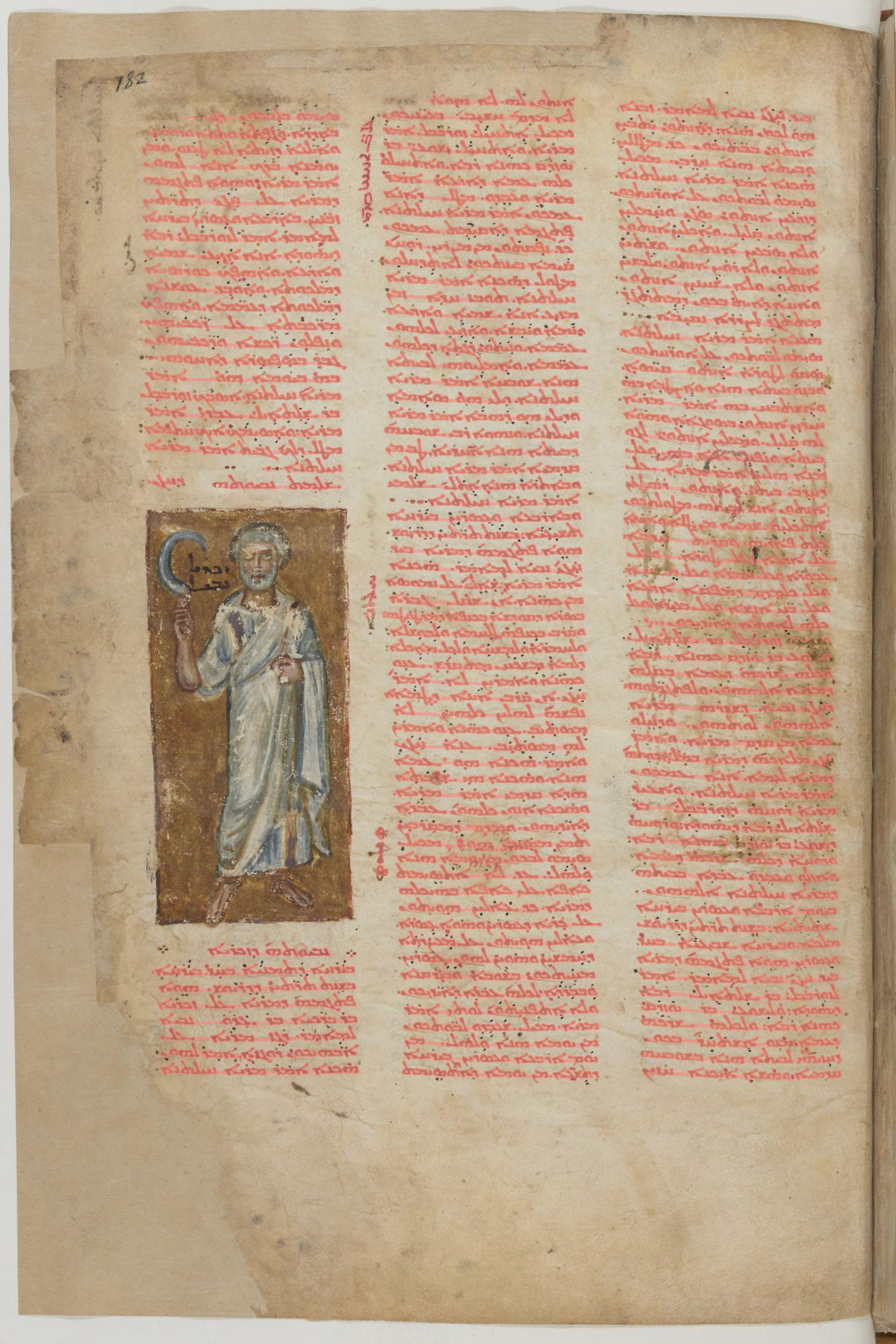}
    \caption{Syriaque 341}
  \end{subfigure}
  
  \vspace{0.5cm}
  
  \begin{subfigure}{0.3\linewidth}
    \includegraphics[width=\linewidth]{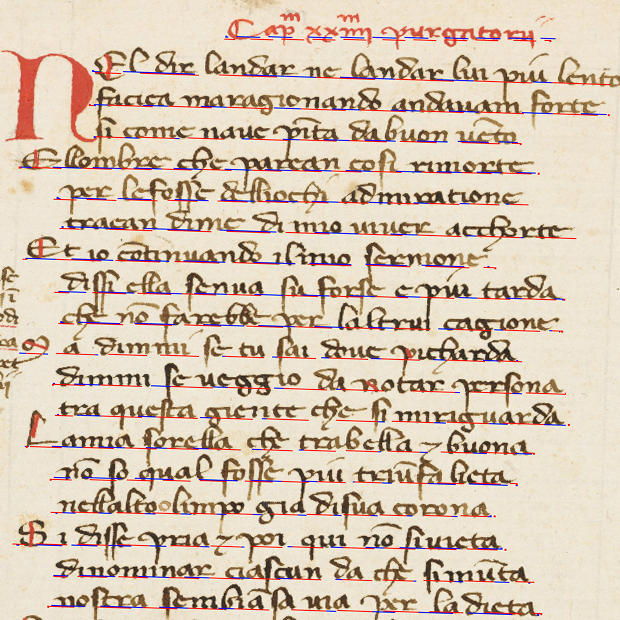}
    \caption{CB55}
  \end{subfigure}
    \begin{subfigure}{0.3\linewidth}
    \includegraphics[width=\linewidth]{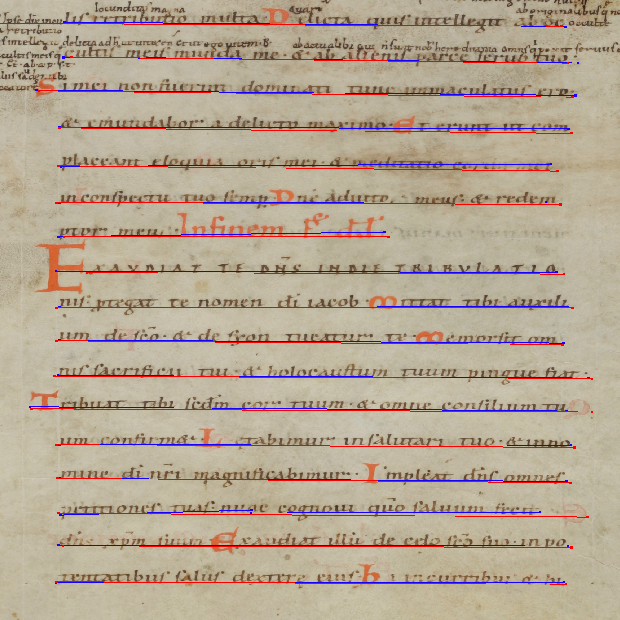}
    \caption{CSG18}
  \end{subfigure}
    \begin{subfigure}{0.3\linewidth}
    \includegraphics[width=\linewidth]{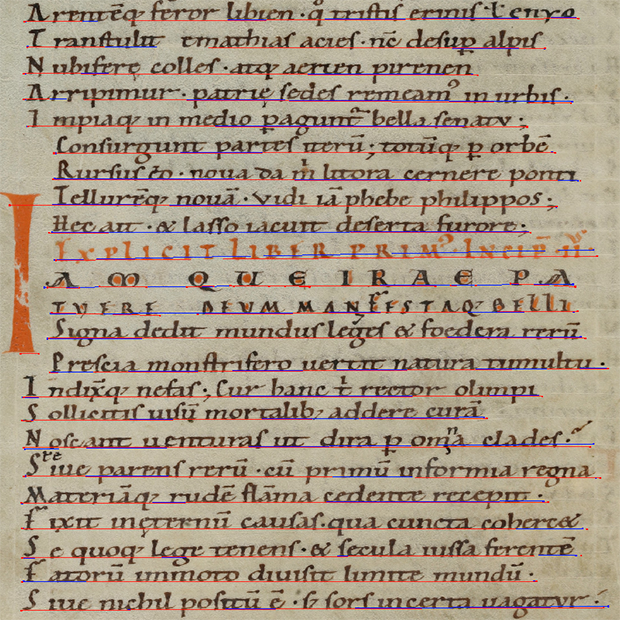}
    \caption{CSG863}
  \end{subfigure}
  \caption{
    Our few-shot method improves SOTA on U-DIADS-TL (Top) by 200\% in RA and 75\% in Line IoU, and matches or surpasses DIVA-HisDB (Bottom) top scores using just three annotated pages.
  }
  \label{fig:qualitative_results}
\end{figure}


\section{Related Work}
In the following, we provide an overview of relevant literature on text line segmentation as well as few-shot learning with an emphasis on its application within historical documents.

\paragraph{Learning-Based Text Line Segmentation.}
Introducing Fully Convolutional Networks (FCNs) shifted research toward an end-to-end, data-driven formulation~\cite{long2015fully}.
UNet~\cite{ronneberger_u-net_2015}, an encoder-decoder architecture with skip connections, demonstrated that precise pixel-level predictions can be learned.
In the context of historical documents, Barakat et al.~\cite{barakat2018text} trained an FCN able to cope with irregular inter-line gaps in medieval Hebrew scrolls. In contrast, Fizaine et al.~\cite{fizaine2024historical} evaluated a Mask-RCNN against UNet on four public archives and found that instance-aware models handle intertwining marginalia more effectively.
More recent transformer hybrids such as SeamFormer~\cite{vadlamudi2023seamformer} employ scribble-conditioned refinement to link diacritics and resolve touching ascenders, achieving SOTA precision on dense palm-leaf manuscripts.
Alberti et al.~\cite{alberti2019labeling} further combined semantic pixel labeling with polygon extraction in a labeling–cutting–grouping pipeline.  

To preserve global structure, recent work augments standard segmentation losses with connectivity-aware terms.  
For instance, such losses have been used for road extraction~\cite{mosinska2018beyond}, for class-imbalance mitigation~\cite{salehi2017tversky}, or have evolved into supervoxel connectivity losses that maintain object continuity with linear complexity and minimal overhead~\cite{grim2025efficient}.

\paragraph{Few-Shot Learning Approaches.}
With the shortage of richly annotated datasets, research shifted to text line segmentation under extreme data scarcity.
Zottin et al.~\cite{zottin2025exploring} explored FCNs, the PSPNet~\cite{zhao2017pspnet}, and DeepLabV3+~\cite{chen_deeplab_2017} with only three labeled pages per manuscript.
In a similar context, De Nardin et al.~\cite{denardin2023efficient} proposed an efficient framework that converts a handful of high-resolution manuscript pages into several thousand training patches via dynamic instance generation and refines coarse masks with Sauvola thresholding, achieving fully supervised quality on DIVA-HisDB with only two annotated pages per manuscript for layout analysis.
An extension replaced Sauvola with adaptive local thresholding while retaining the same pixel-level F-score under a stricter evaluation \cite{denardin2023few_shot}.
Follow-up work demonstrated that a single labeled page per Arabic manuscript can suffice when combined with lightweight augmentation and class-balanced losses \cite{denardin2024one_shot}.  

\section{Methodology}
In the following section, we describe our methodology to enable efficient few-shot learning capabilities for text line segmentation and baseline detection of ancient manuscripts.

\subsection{Model Architecture}
We adopt UNet++~\cite{zhou2018unetplusplus}, derived from the original UNet~\cite{ronneberger_u-net_2015}, paired with a ResNet34~\cite{he2016deep} encoder.
Despite its simplicity, the UNet++ consistently outperforms more complex architectures, such as the commonly used DeepLabV3+ architecture~\cite{zottin2025exploring,denardin2023efficient}.
We attribute this to better generalization, reduced overfitting, and suitability to the data-scarce nature of historical document segmentation.

In comparison to \cite{zottin2025exploring}, we opt for training on patch-level as opposed to the whole image, as this has been an established practice in domains where data is scarce, such as in the case of historical documents \cite{denardin2023efficient,sterzinger2024drawing}.
With accompanying benefits such as better capturing fine-grained text line structures or reducing memory usage, we apply our model to random crops extracted from high-resolution input images during training and overlapping patches during validation and testing.
In detail, we extract patches of size $448\times448$ pixels and perform random rotations between -5 and +5 degrees and a random shear transformation between -3 and +3 degrees. 
We incorporate these augmentations to include non-horizontally aligned text lines.

To obtain a prediction for a whole page, overlapping patches are first processed independently, and predictions are stitched back together to form the full-size image.
To avoid visual artifacts at border regions, each patch is scaled by a 2D Gaussian function, emphasizing its central region, which ensures smooth and seamless blending across overlapping areas.

\begin{figure}[t]
    \centering
    \begin{subfigure}[]{0.24\linewidth}
        \includegraphics[width=\linewidth]{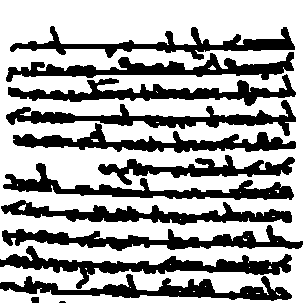}
        \caption{Ground Truth}
        \label{fig:gt_sample}
    \end{subfigure}
    \begin{subfigure}[]{0.24\linewidth}
        \includegraphics[width=\linewidth]{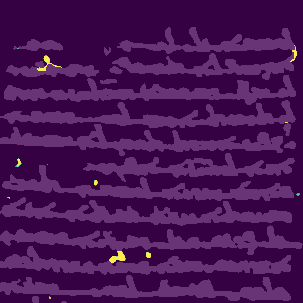}
        \caption{$\alpha=1.0,\beta=0.0$}
    \end{subfigure}
     \begin{subfigure}[]{0.24\linewidth}
        \includegraphics[width=\linewidth]{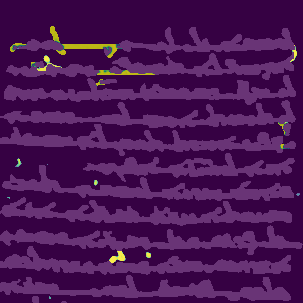}
        \caption{$\alpha=1.0,\beta=0.5$}
    \end{subfigure}
    \begin{subfigure}[]{0.24\linewidth}
        \includegraphics[width=\linewidth]{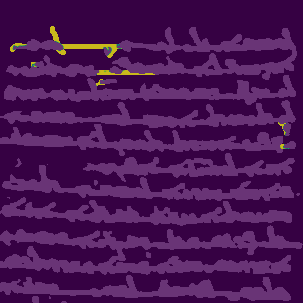}
        \caption{$\alpha=1.0,\beta=1.0$}
    \end{subfigure}
    \vspace{0.5cm}
    \begin{subfigure}[]{0.24\linewidth}
        \includegraphics[width=\linewidth]{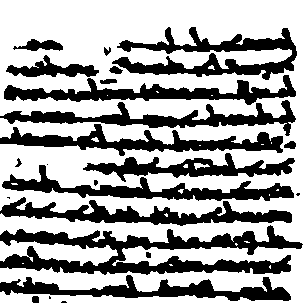}
        \caption{Prediction}
    \end{subfigure}
    \begin{subfigure}[]{0.24\linewidth}
        \includegraphics[width=\linewidth]{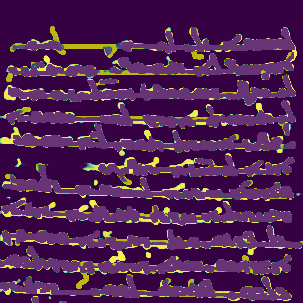}
        \caption{$\alpha=0.5,\beta=0.0$}
    \end{subfigure}
    \begin{subfigure}[]{0.24\linewidth}
        \includegraphics[width=\linewidth]{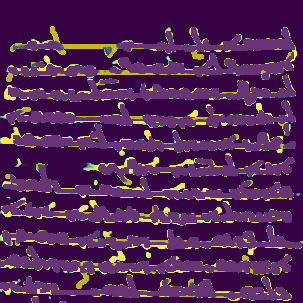}
        \caption{$\alpha=0.5,\beta=0.5$}
    \end{subfigure}
    \begin{subfigure}[]{0.24\linewidth}
        \includegraphics[width=\linewidth]{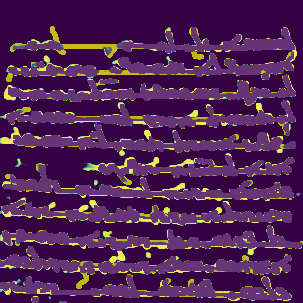}
        \caption{$\alpha=0.5,\beta=1.0$}
    \end{subfigure}      
    
    \caption{An illustration of the parameters $\alpha$ and $\beta$ and their effect on the loss using a sample patch of Syriaque 341: they control the trade-off between pixel- and structure-level errors, and between split and merge penalties, respectively~\cite{grim2025efficient}.}
    \label{fig:structure_loss}
\end{figure}
\subsection{Connectivity-Preserving Loss}
As text lines, by their very nature, are continuous, curvilinear structures (cf. \Cref{fig:gt_sample}), and maintaining their integrity -- avoiding fragmentation or unintended merging -- is crucial for accurate text line segmentation and detection.
Concerning commonly used evaluation metrics for text line segmentation, minor errors at a pixel level, lead to significant drops in performance at the line-level.

We therefore propose the integration and adaptation of a novel connectivity-aware loss function, initially introduced by Grim et al.~\cite{grim2025efficient} for neuron morphology reconstruction.
In detail, this loss also penalizes topological errors at a global component level rather than solely at the pixel level.

Formally, the loss function is expressed as:

\begin{align}
    \mathcal{L}(y,\hat{y}) &= (1-\alpha)\mathcal{L}_{0}(y,\hat{y}) + \alpha\Big((1-\beta)\sum_{C\in\mathcal{N}(\hat{y}_{-})}\mathcal{L}_{0}(y_{C},\hat{y}_{C}) \\&+ \beta\sum_{C\in\mathcal{P}(\hat{y}_{+})}\mathcal{L}_{0}(y_{C},\hat{y}_{C})\Big).
\end{align}

Here, $y$ represents the ground truth text line segmentation and $\hat{y}$ is the network's prediction.
Furthermore, the sets $\mathcal{N}(\hat{y}_{-})$ and $\mathcal{P}(\hat{y}_{+})$ identify components that would alter the number of connected components in the predicted text lines.
It is important to note that in our implementation, we use an \textit{inverted ground truth} representation.
Consequently, $\mathcal{P}(\hat{y}_{+})$ captures false negative regions that cause a text line to appear fragmented (a false split), while $\mathcal{N}(\hat{y}_{-})$ captures false positive regions that bridge two distinct ground truth lines (a false merge).
Moreover, $\mathcal{L}_0$ refers to a standard pixel-level objective, such as, e.g., the binary cross-entropy loss.
In our methodology, we adapted the loss by adding the Dice loss~\cite{sudre2017generalised} and weighing the two terms to be of the same magnitude:
\begin{equation}
1 - \frac{2 \sum_i^N y_i \hat{y}_i}{\sum_i^N y_i + \sum_i^N \hat{y}_i} + \frac{10}{N}\sum_i^N\mathcal{L}(y_i,\hat{y_i}).
\end{equation}

Finally, the hyperparameters $\alpha$ and $\beta$ provide crucial control over the connectivity: $\alpha \in [0,1]$ balances the influence of the conventional pixel-level loss against the topological penalties, with higher values prioritizing the correction of connectivity errors.
As this is vital for maintaining the precise structure of text lines in ancient manuscripts prone to fragmentation or merging, we select $\alpha=1$.
Concurrently, $\beta \in [0,1]$ allows for differential weighting between false merge and false split errors in text line segmentation; a $\beta < 0.5$ prioritizes penalizing merged lines, while a $\beta > 0.5$ emphasizes correcting fragmented lines, enabling tailored error correction based on the specific challenges of the segmentation task.
Interestingly, for both our selected datasets, we found that focusing exclusively on penalizing false merged lines yielded the best performance, and thus we chose $\beta=0$.
To better illustrate the impact of these hyperparameters, \Cref{fig:structure_loss} shows the penalized error in yellow overlaid on the prediction.

Due to the high computational cost of calculating the connected components and matching them, we utilize the structure-aware loss solely to fine-tune a model that has been pretrained with only the Dice loss.
For similar reasons, we do not employ the structure-aware loss for model selection during the fine-tuning stage. Instead, we utilize the mean IoU as a proxy function to evaluate performance during validation.

\begin{figure}
  \centering
  \includegraphics[width=0.9\linewidth]{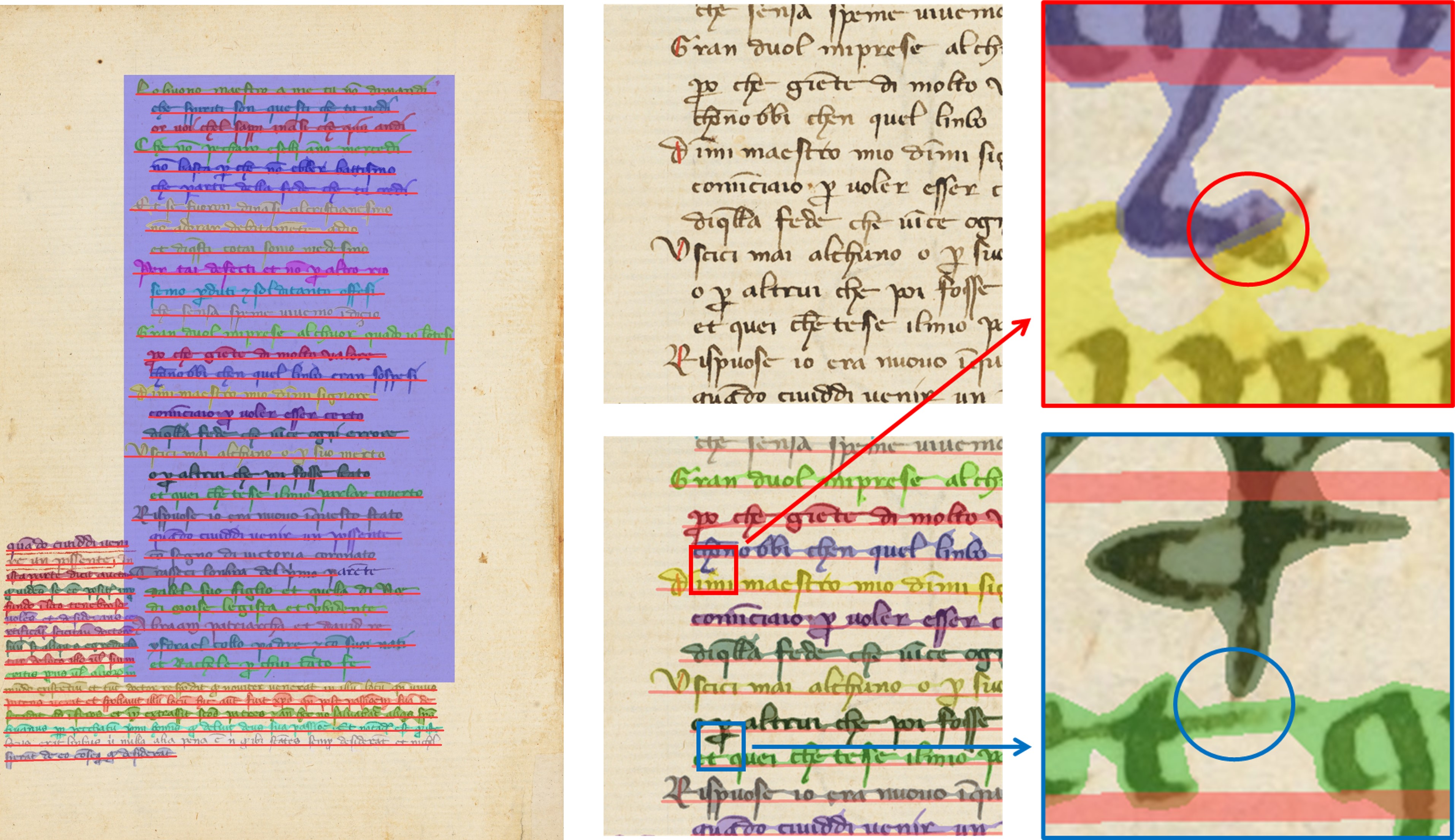}
  \caption{
    Example excerpts of the CB55 manuscript in DIVA-HisDB illustrating the issue of overlapping text line segments.
  }
  \label{fig:diva-detail}
\end{figure}

\section{Datasets}
\label{sec:datasets}

Our experiments are grounded in publicly available datasets from recent academic competitions focused on advancing text line segmentation for historical documents.
These benchmarks, including the ICDAR 2017 HisDoc-Layout-Comp~\cite{simistira2017compe} and the ICDAR 2025 FEST track~\cite{zottin2024icdar}, emphasize challenges such as complex layouts, document degradation, and the latter, critically, few-shot learning constraints where only a handful of annotated examples are available for training.
We use two such datasets: U-DIADS-TL\cite{zottin2025exploring,icdar2025fest} and DIVA-HisDB~\cite{simistira2017compe}, a widely-used dataset for historical layout analysis.
While our primary development targets the proposed few-shot paradigm of U-DIADS-TL, we also adapt and evaluate DIVA-HisDB in a few-shot setting to test the generalization of our method.

\subsection{U-DIADS-TL}

The U-DIADS-TL dataset is the official benchmark for the FEST 2025 competition and is explicitly designed for few-shot text line segmentation~\cite{zottin2025exploring}. It is a subset of the larger U-DIADS-Bib~\cite{zottin2024udiads} collection, which provides a rich resource for multi-class layout analysis.

\paragraph{Technical Details.}
The dataset contains 105 high-resolution color images from three distinct medieval manuscripts: Latin 2, Latin 14396, and Syriaque 341. These sources feature diverse scripts, two- and three-column layouts, interlinear glosses, and significant degradation. For each manuscript, the data is strictly partitioned into 3 pages for training, 10 for validation, and 22 for testing, thus requiring effective methods to handle such low data regimes.
The ground truth consists of pixel-precise binary masks that distinguish text lines from the background, with no overlap between line instances.
Our methodology is designed to adhere to the official evaluation protocol described in \cite{zottin2025exploring,icdar2025fest}.

\subsection{DIVA-HisDB}
The DIVA-HisDB dataset was introduced for the ICDAR 2017 HisDoc-Layout-Comp~\cite{simistira2017compe} competition.
It is recognized for its complex page layouts, which include marginalia, decorated initials, and interlinear glosses.

\paragraph{Technical Details.}
The DIVA-HisDB consists of 150 pages from three manuscripts (CB55, CSG18, CSG863), provided as 600 dpi RGB scans with ground truth in PAGE XML format.
For our experiments, we downsample all images by a factor of three to make the image dimensions between the two datasets coherent.
The dataset supports multiple layout analysis tasks, including full text line segmentation (Task 3), equal to the FEST 2025 competition.
However, the polygon annotations in PAGE XML format for this task feature overlapping regions between adjacent lines, as illustrated in \Cref{fig:diva-detail}.
This characteristic is incompatible with models designed to produce a single, non-overlapping binary segmentation mask, making a pixel-level evaluation ambiguous.

To address this, we opt for the task of baseline detection (Task 2), which requires the prediction of start and end-points of baselines instead of full polygons, providing a suitable proxy for text line localization without the issue of overlapping regions.
To maintain consistency with the few-shot evaluation setting of U-DIADS-TL, we do not use the full training partition of 20 pages per manuscript, but instead, for each experiment, we randomly sample three pages for training, allowing us to assess our model's few-shot generalization.

\section{Evaluation}

In the following section, we evaluate our proposed methodology. We begin by introducing the relevant evaluation metrics and experimental setup, followed by an ablation study of each component using our primary dataset, U-DIADS-TL. Finally, we assess the full approach on both benchmark datasets to underscore its effectiveness and generalizability.

\subsection{Metrics}

For the U-DIADS-TL dataset, we report Pixel IoU and Line IoU as well as Detection Rate (DR), Recognition Accuracy (RA), and F-Measure (FM) from matched line segments as proposed by Zottin et al.~\cite{zottin2025exploring,icdar2025fest}.
Pixel IoU and Line IoU are metrics based on the IoU, which is defined as:

\begin{equation}
\text{IoU} = \frac{TP}{TP + FP + FN},
\label{eq:iou}
\end{equation}

where $TP$ is the number of true positives, $FP$ false positives, and $FN$ false negatives of the predicted binary mask.
Subsequently, Pixel IoU evaluates segmentation quality at the pixel level, whereas Line IoU evaluates detection accuracy at the line level, where connected components are matched if pixel-level precision and recall exceed a threshold of 75\%.

Concerning DR, RA, and FM, these metrics are based on a matching score, MatchScore.
For predicted line $R_i$ and ground-truth line $G_j$, the MatchScore is defined as:
\begin{equation}
\text{MatchScore}(i,j) = \frac{T(G_j \cap R_i)}{T(G_j \cup R_i)},
\label{eq:matchscore}
\end{equation}
where a pair $(i,j)$ is considered a one-to-one match if $\text{MatchScore}(i,j) \geq 0.75$.
Let $M$ be the number of such matches, $N_1$ the number of ground-truth lines, and $N_2$ the number of detected lines, then the match-based metrics are defined as follows:
\begin{equation}
\text{DR} = \frac{M}{N_1},\ \text{RA} = \frac{M}{N_2},\ \text{FM} = \frac{2 \cdot \text{DR} \cdot \text{RA}}{\text{DR} + \text{RA}}. \label{eq:metrics}
\end{equation}

Based on overlapping text line annotations and resulting problems when evaluating text line segmentation on pixel-level (cf. \Cref{fig:diva-detail}), for the DIVA-HisDB dataset, we opt for the simplified task of text baseline detection.
With this comes the necessity of employing suitable evaluation metrics for which we follow closely the evaluation protocol denoted in the work by Simistira et al.~\cite{simistira2017compe}, which is based on the cBAD competition~\cite{diem2017cbad}.
Here, an F-Measure metric is employed, which assesses baseline extraction by calculating recall and precision for polygonal chains representing baselines~\cite{diem2017cbad}.

\subsection{Evaluation Protocol}
For model training, an AdamW optimizer is employed with a learning rate of $3e-4$, and models are trained or fine-tuned for 100 epochs using a batch size of eight.
An early stopping criterion is implemented to prevent overfitting, halting training after 10 iterations and fine-tuning after 30 iterations without improvement in validation performance.
Model selection for final evaluation is based on the highest validation mean IoU achieved during each run.

Regarding evaluation across datasets, we report the average performance obtained from training with at least three random seeds for U-DIADS-TL.
Conversely, for the DIVA-HisDB dataset, performance is reported from the model that achieves the highest validation performance, consistent with its competition setting.
Furthermore, specific to the DIVA-HisDB dataset, a post-processing pipeline is applied to facilitate comparison with ICDAR competition results from 2017~\cite{simistira2017compe}, which include filtering out segments shorter than 50 pixels and subsequently merging lines that are both geometrically close (within a 50-pixel distance threshold) and similarly oriented (within a 15-degree angle threshold).

\begin{table}
    \centering

     \begin{tabular}{l >{\centering\arraybackslash}p{2cm} >{\centering\arraybackslash}p{1.5cm} >{\centering\arraybackslash}p{1.5cm} >{\centering\arraybackslash}p{1cm} >{\centering\arraybackslash}p{1cm}}
\toprule
        \textbf{Name} &  \textbf{Pixel IoU} &  \textbf{Line IoU} &    \textbf{DR} & \textbf{RA} & \textbf{FM} \\
\midrule
Baseline~\cite{zottin2025exploring} & $49.0$ & $44.3$ & $29.1$ & $17.7$ & $21.7$ \\
DeepLabV3+  & 52.44 & 51.90 & 48.32 & 30.00 & 36.44 \\
\midrule
+ UNet++ & 64.50 & 70.06 & 66.97 & 41.83 & 50.77 \\
+ Lighter Encoder & 65.29 & 71.24 & 68.30 & 42.39 & 51.94 \\
+ Connectivity Loss & 67.83 & 75.17 & 72.07 & 46.36 & 55.66 \\
\midrule
\textbf{Final Model (Tuned)} & \textbf{70.64} & \textbf{77.38} & \textbf{73.90} & \textbf{52.96} & \textbf{61.40} \\
\bottomrule
\end{tabular}
    \caption{Ablation Study of the Proposed Method.}
    \label{tab:ablation_study}
\end{table}

\subsection{Ablation}

In the following section, we will ground our proposed method by conducting an ablation study on the individual components. 
Each component yields significant performance improvements, which are denoted in \Cref{tab:ablation_study}, which summarizes the results of the performed study.

\subsubsection{Model Architectures}
First, we ablate different model architectures.
For architecture selection, we considered the same as Zottin et al.~\cite{zottin2025exploring} plus a transformer-based one; however, this time, training on patches instead of the whole image this time.
Inspecting the results in \Cref{tab:ablation_architectures}, we observe that simpler architectures, e.g., the UNet~\cite{ronneberger_u-net_2015} or derived variants, i.e., the UNet++~\cite{zhou2018unetplusplus}, yield drastically better performance over all metrics.
In line with the theory that less bias reduces the risk of overfitting~\cite{goodfellow2016deep}, we select the UNet++ given our context of data-scarce training regimes in the few-shot setting.

\begin{table}
    \centering
     \begin{tabular}{l >{\centering\arraybackslash}p{2cm} >{\centering\arraybackslash}p{1.5cm} >{\centering\arraybackslash}p{1cm} >{\centering\arraybackslash}p{1cm} >{\centering\arraybackslash}p{1cm}}
\toprule
        \textbf{Name} & \textbf{Pixel IoU} & \textbf{Line IoU} & \textbf{DR} & \textbf{RA} &    \textbf{FM} \\
\midrule
     PSPNet\cite{zhao2017pspnet} &     42.22 & 44.08 & 37.31 & 33.33 & 34.18 \\
     SegFormer~\cite{xie_segformer_2021} &     50.10 & 49.27 & 46.42 & 29.66 & 35.50 \\
     UNet++~\cite{zhou2018unetplusplus} &     64.50 & 70.06 & 66.97 & 41.83 & 50.77 \\
     DeepLabV3+~\cite{chen_deeplab_2017} &     52.44 & 51.90 & 48.32 & 30.00 & 36.44 \\
     UNet~\cite{ronneberger_u-net_2015} & 64.34 & 69.56 & 66.71 & 39.92 & 49.31 \\
\bottomrule
\end{tabular}
    \caption{Ablating Architectures.}
    \label{tab:ablation_architectures}
\end{table}

\vspace{-1cm}

\subsubsection{Encoder Parameter Size}
\begin{figure}
    \centering
    \includegraphics[width=0.94\linewidth]{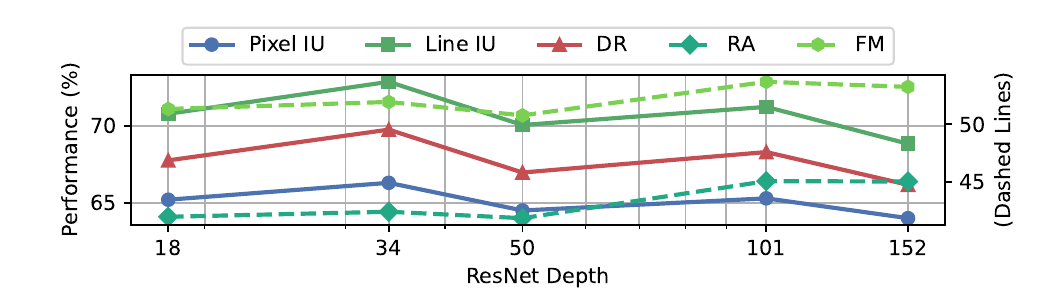}
    \caption{Ablation on Network Depth.}
    \label{fig:ablation_depth}
\end{figure}

Continuing with the previous made insight, we ablate encoder depth, evaluating ResNets~\cite{he2016deep} pre-trained on ImageNet~\cite{de2023imagenet} with varying amounts of network layers.
We illustrate our findings in \Cref{fig:ablation_depth}, which indicates a similar trend: performance tends to increase with decreasing depth.
However, this trend is not stable as there is some fluctuation.
For instance, a ResNet with a depth of 18 performs clearly worse than one with 34 layers, while being better than larger variants with up to 152 layers.

\subsubsection{Connectivity Loss}
An essential component of our proposed methodology for few-shot text line segmentation is the tuned connectivity-aware loss function proposed by Grim et al.~\cite{grim2025efficient} and leveraged within our methodology.
As the hyperparameters $\alpha$ and $\beta$ influence the trade-off between pixel-level and topological loss and between penalizing false merge and false split errors, we perform an ablation study on these parameters to select them optimally.

\begin{figure}[h]
    \centering
    \begin{subfigure}{\linewidth}
        \includegraphics[width=0.94\linewidth]{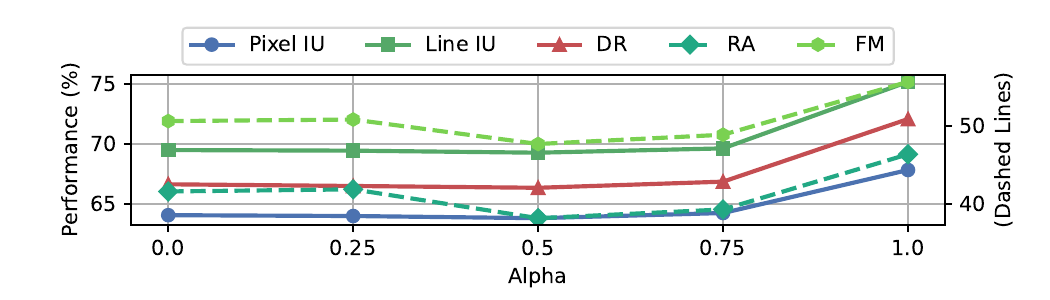}
        \caption{Varying $\alpha$, while keeping $\beta$ constant at 0.5.} 
        \label{fig:alpha}
    \end{subfigure}
    \begin{subfigure}{\linewidth}
        \includegraphics[width=0.94\linewidth]{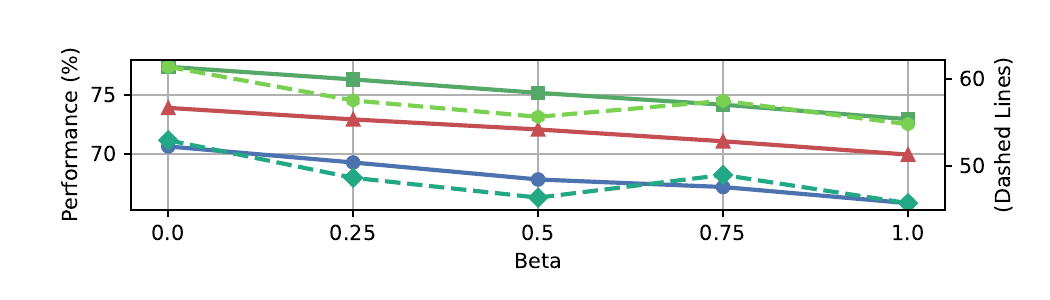}
        \caption{Varying $\beta$, while keeping $\alpha$ constant at 1.} 
        \label{fig:beta}
    \end{subfigure}
    \caption{Ablating the Hyper-Parameters of the Connectivity Loss.}
    \label{fig:hyper_ablation}
\end{figure}

We illustrate our findings in \Cref{fig:hyper_ablation} and observe in \Cref{fig:alpha} that performance is best when considering only the topology of the mask (i.e., $\alpha=1$), neglecting pixel-level performance, which is expected due to our additional incorporation of the Dice loss.
In contrast to initial assumptions that both false merges and false splits play a crucial role in performance, results of \Cref{fig:beta} show that only false merges are problematic (i.e., $\beta=0$).
Given these results, we performed the same ablation for the DIVA-HisDB baseline detection task and found $\beta=0$ to also work best.

\subsection{Results}

Our comprehensive evaluation on two distinct historical document datasets, DIVA-HisDB and U-DIADS-TL, unequivocally demonstrates the competitive performance and generalizability of our proposed few-shot methodology in extremely scarce data settings of only three available annotated pages. Summarized in \Cref{tab:final_diva} and \Cref{tab:final_udiads}, our results highlight the impact of the connectivity-aware loss, which we propose for text line segmentation and baseline detection, in addition to our use of smaller and simpler architectures compared to previous SOTA methods.

\begin{table}[h]
    \centering

     \begin{tabular}{l >{\centering\arraybackslash}p{1.5cm} >{\centering\arraybackslash}p{1.5cm} >{\centering\arraybackslash}p{1.5cm} >{\centering\arraybackslash}p{1.5cm}
     >{\centering\arraybackslash}p{1.5cm}}
\toprule
        \textbf{Name} &  \textbf{Samples} & CB55 & CSG18 & CSG863 & \textbf{Avg. F1} \\
\midrule
System-8~\cite{simistira2017compe} & \multirow{3}{*}{20} & 98.96 & 98.53 & 97.16 & \textbf{98.22} \\
System-2~\cite{simistira2017compe} & & 95.97 & \textbf{98.79} & \textbf{98.30} & 97.68 \\
UNet++ & & \textbf{99.66} & 94.31 & 92.24 & 95.40 \\
\midrule
UNet++ & \multirow{3}{*}{\underline{3}} & 93.17 & 83.17 & 86.62 & 87.65 \\
+Connectivity Loss & & 96.07 & 87.27 & 86.02 & 89.79 \\ 
+Post-Processing &  & \underline{99.00} & \underline{94.13} & \underline{93.62} & \underline{95.58} \\
\bottomrule
\end{tabular}

    \caption{Final Results DIVA-HisDB.}
    \label{tab:final_diva}
\end{table}

In more detail, on the DIVA-HisDB dataset with the task of baseline detection (\Cref{tab:final_diva}), our suggested lightweight UNet-based methodology, trained on patches with only three annotated pages, achieves an average F1 score of 95.58\%.
In comparison to the UNet++ baseline trained on a significantly larger dataset (20 samples versus 3, 95.40\% F1) and approaching or even outperforming the F1 scores of top competition systems (System-8: 98.22\%, System-2: 97.68\%) that leveraged considerably more training data, the efficacy of our methodology in the few-shot setting is underscored. 

\begin{table}[h]
\centering
\begin{tabular}{l >{\centering\arraybackslash}p{1.5cm}>{\centering\arraybackslash}p{1cm}>{\centering\arraybackslash}p{1.5cm}|>{\centering\arraybackslash}p{1.5cm} >{\centering\arraybackslash}p{1.5cm}|>{\centering\arraybackslash}p{2cm}}
\toprule
\textbf{Metric} & \textbf{Latin 14396} & \textbf{Latin 2} & \textbf{Syriaque 341} & \textbf{Avg. Ours} & \textbf{Avg. \cite{zottin2025exploring}} & \textbf{Rel. Gain to \cite{zottin2025exploring}} \\
\midrule
Pixel IoU             & 73.71 & 77.83 & 60.36 & 70.64 & 49.0 & $\ +44.16$\\
Line IoU              & 79.78 & 86.28 & 66.09 & 77.38 & 44.3 & $\ +74.67$\\
DR & 79.45 & 84.37 & 57.89 & 73.90 & 29.1 & +153.95\\
RA & 51.20 & 65.40 & 42.27 & 52.96 & 17.7 & +199.21\\
F-Measure            & 61.96 & 73.43 & 48.83 & 61.40 & 21.7 & +182.95\\
\bottomrule
\end{tabular}
\caption{Final Results U-DIADS-TL.}
\label{tab:final_udiads}
\end{table}

On the more challenging task of text line segmentation of the U-DIADS-TL dataset (Table \ref{tab:final_udiads}), the impact of our methodology is further exemplified, where our final, tuned model sets a new SOTA for few-shot text line segmentation.
Across all metrics, our approach achieves relative gains ranging from 75\% to 200\%.
Specifically, we observe a nearly 200\% gain in RA and an approximate 185\% gain in FM. 
Complementary improvements include a plus of around 75\% in Line IoU (reaching 77.38\%), an over 150\% relative gain in DR (attaining 73.90\%), and a nearly 45\% relative improvement in Pixel IoU, with an absolute performance of 70.64\%.
An excerpt of qualitative results is depicted in \Cref{fig:qualitative_results}.

\section{Conclusion}
In this work, we challenged the prevailing trend of increasing model complexity for few-shot text line segmentation in historical documents.
We demonstrated that a simpler, lightweight UNet++ architecture, when paired with a connectivity-aware loss function, yielded superior performance in a low-data regime.
Our approach directly addressed the critical issue of maintaining topological correctness by penalizing common structural errors such as line fragmentation and inadvertent merges.

Our experimental results substantiated the effectiveness of this methodology.
On the U-DIADS-TL benchmark, our model established a new state-of-the-art, improving the Line IoU by over 74\% and the RA by nearly 200\% relative to the baseline.
Furthermore, it showed strong generalization capabilities on the DIVA-HisDB dataset, producing results competitive with top-performing systems while utilizing 85\% less training data.
These findings confirmed our central hypothesis: architectural simplicity, when combined with a topologically informed loss function, provides a more potent and efficient solution for few-shot segmentation than more complex, data-hungry models.

\bibliographystyle{splncs04}
\bibliography{bibliography}

\end{document}